\documentclass[a4paper]{llncs}

\usepackage{graphicx}
\usepackage{amsmath,amssymb}
\usepackage{mathrsfs}
\usepackage{subfigure}
\usepackage{url}

\renewcommand{\emptyset}{\varnothing}
\renewcommand{\leq}{\leqslant}

\newcommand{\VarDomain}[1]{\ensuremath{D\lbrack#1\rbrack}}
\newcommand{\CtrScope}[1]{\ensuremath{S\lbrack#1\rbrack}}
\newcommand{\CtrProjection}[3]{\ensuremath{\Pi(#1,#3,#2)}}
\newcommand{\NetSetVar}{V}
\newcommand{\NetSetCtr}{R}
\newcommand{\NetSetDom}{D}
\newcommand{\Network}{P}
\newcommand{\Interval}[2]{\ensuremath{\lbrack#1,#2\rbrack}}
\newcommand{\DBC}{DBC}

\newcommand{\AlgoName}[1]{\textsf{#1}}
\newcommand{\AlgoKey}[1]{\textbf{#1}}
\newcounter{AlgoLine}
\newcommand{\AlgoNumSet}{\setcounter{AlgoLine}{1}}
\newcommand{\AlgoNum}{{\tiny\arabic{AlgoLine}.}\stepcounter{AlgoLine}}

\newcommand{\GSet}[1]{\ensuremath{\mathscr{#1}}}
\newcommand{\ISet}{\ensuremath{\mathbb{I}}}
\newcommand{\RSet}{\ensuremath{\mathbb{R}}}
\newcommand{\NewMathOp}[2]{\ensuremath{\mathrm{\mathop{#1}}(#2)}}
\newcommand{\Hull}[1]{\NewMathOp{hull}{#1}}
\newcommand{\ReviseBounds}[2]{\NewMathOp{ReviseBounds}{#1,#2}}

\newcommand{\ReviseBoundsZA}{\ensuremath{\mathrm{\mathop{ReviseBounds}}}}

\title{Directional Consistency for \\
  Continuous Numerical Constraints}

\author{Fr\'ed\'eric Goualard \and Laurent Granvilliers}

\institute{LINA -- University of Nantes -- France\\
2, rue de la Houssini\`ere -- BP 92208 -- F-44322 Nantes cedex 3\\
\{Frederic.Goualard$\mid$Laurent.Granvilliers\}@lina.univ-nantes.fr}

\begin{document}
\maketitle

\begin{abstract}
  Bounds consistency is usually enforced on continuous constraints by
  first decomposing them into binary and ternary primitives. This
  decomposition has long been shown to drastically slow down the
  computation of solutions.  To tackle this, Benhamou et al.\ have
  introduced an algorithm that avoids formally decomposing
  constraints.  Its better efficiency compared to the former method
  has already been experimentally demonstrated.  It is shown here that
  their algorithm implements a strategy to enforce on a continuous
  constraint a consistency akin to Directional Bounds Consistency as
  introduced by Dechter and Pearl for discrete problems. The algorithm
  is analyzed in this framework, and compared with algorithms that
  enforce bounds consistency.  These theoretical results are
  eventually contrasted with new experimental results on standard
  benchmarks from the interval constraint community.
\end{abstract}

\section{Introduction}\label{sec:introduction}

Waltz's seminal paper~\cite{Waltz:75} promoted the idea of local
consistency enforcement to solve constraints. Systems of constraints
were solved by considering each of them in turn, discarding the values
in the domains of the variables involved that could not possibly be
part of a solution. Montanari~\cite{Montanari:IS74} and
Mackworth~\cite{Mackworth:AI77} introduced the notion of a network of
constraints in which a more involved scheme for propagating domain
modifications could be used. Davis~\cite{Davis:AI1987} later adapted
these works to solve continuous problems by employing interval
arithmetic~\cite{Moore:66} to handle the domains of the variables.

The first solvers to implement the ideas of Davis and others were
enforcing on continuous constraints a relaxation of arc consistency,
\emph{bounds consistency} (aka 2B consistency~\cite{Lhomme:IJCAI93}),
which is better suited to real domains.  For practical reasons, bounds
consistency can only be effectively enforced on binary and ternary
constraints.  More complex constraints then have to be decomposed into
such simpler constraints, thereby augmenting the number of variables
and constraints to eventually consider.  Benhamou, van Hentenryck, and
McAllester~\cite{Benhamou-et-al:ILPS94} produced experimental
evidences that such a process drastically slows down the computation,
rendering in effect bounds consistency computation impracticable on
many problems. They then advocated to replace bounds consistency
altogether by a new consistency notion, \emph{box consistency}, whose
enforcement does not require the decomposition of complex constraints.

However, in 1999, Benhamou et al.\ presented the HC4
algorithm\footnote{It has come recently to our attention that an
  algorithm equivalent to HC4 was independently discovered by
  Messine~\cite{Messine:PhD1997}. To our knowledge, this author did
  not study its theoretical properties,
  though.}~\cite{Benhamou:ICLP1999}, which is strongly related to the
methods employed to enforce bounds consistency except for its ability
to handle constraints without formal decomposition.  HC4 was shown to
outperform box consistency-based solvers on some large problems; in
addition, its use in a suitable cooperation scheme was recommended to
speed up the computation of box consistency on difficult problems. In
their paper, the authors did not analyze HC4 on a theoretical
point-of-view.  They claimed, however, that it would enforce bounds
consistency on the system of primitive constraints stemming from the
decomposition of a constraint containing no more than one occurrence
of any variable.

The contribution of this paper is to present an analysis of the HC4
algorithm and to compare it from the theoretical point-of-view with
the HC3 algorithm used to enforce bounds consistency on decomposed
constraints. We also characterize the consistency HC4 enforces on one
constraint in terms of the equivalent on continuous domains of
\emph{directional bounds consistency} introduced by Dechter and
Pearl~\cite{Dechter-Pearl:AI87}, and we prove Benhamou et al.'s claim
concerning it computing bounds consistency for constraints with
variables occurring at most once. Lastly, we analyze experimental results
to justify the discrepancy they exhibit with theoretical results.

To offer a reasonably self-content exposition, we start by recalling
some definitions and algorithms related to the solving of discrete
Constraint Satisfaction Problems (CSP) in
Section~\ref{sec:discrete-csp}; We then adapt in
Section~\ref{sec:directional-bounds-consistency} the framework just
introduced to the case of continuous CSPs, thus emphasizing the
parallel between algorithms presented by
Dechter~\cite{Dechter:book2003} to enforce Directional Arc Consistency
and the HC4 algorithm; The complexity of HC4 and HC3 are compared in
Section~\ref{sec:theoretical-analysis}, first for one constraint only
in Section~\ref{sec:complexity-one-constraint}, and then on a
constraint system in Section~\ref{sec:complexity-constraint-system};
in Section~\ref{sec:experimental-results}, we contrast the theoretical
results obtained in the previous sections with experimental results
obtained on a set of standard benchmarks; finally, we analyze our
results in Section~\ref{sec:discussion}, and we propose some other
interpretations of the HC4 algorithm that could lead to more efficient
algorithms.

\section{Local Consistency Techniques for Discrete Problems}\label{sec:discrete-csp}

The entities we will manipulate in the sequel of this paper are
\emph{variables} from an infinite countable set $\{x_1,x_2,\dots\}$,
with their associated domains of possible values $\VarDomain{x_1}$,
$\VarDomain{x_2}$, \dots, and \emph{constraints}, which enforce some
relation between variables.  A \emph{constraint} $C$ on a finite set
of variables $\CtrScope{C}$---its \emph{scope}---with domains
$D$ is a subset of the product of their domains.

A \emph{constraint problem} $\Network$ is formally characterized by a triplet
$(\NetSetVar,\NetSetDom,\NetSetCtr)$, where $\NetSetVar$ is a finite
set of variables with domains $\NetSetDom$, and $\NetSetCtr$ is a set
of constraints such that the union of their scopes is included in 
$\NetSetVar$.

The basic step to solve a
constraint problem corresponds to the inspection of each of its
constraints $C$ in turn and to the removal of all the values in the domains
of the variables in \CtrScope{C} that cannot be part of a solution of $C$. To this end, 
we need to be able to project $C$ onto each of its variables. This notion of
\emph{projection} is formally stated below.

\begin{definition}[Projection of a constraint]
  Let $C$ be a constraint, $D$ a Cartesian product of domains, and
  $x$ a variable in $C$. The \emph{projection}
  of $C$ on $x$ w.r.t.\ $D$ is the set $\CtrProjection{C}{x}{D}$ of values in
  $\VarDomain{x}$ that can be extended to a solution of $C$ in $D$.
\end{definition}

A projection is then the set of \emph{consistent} values of a variable
relative to a constraint. The well-known \emph{arc
  consistency} property~\cite{Mackworth:AI77} demands that all values
in a domain be consistent.  This property is clearly too strong for
our purpose since it has no practical counterpart for continuous
domains in the general case. We will then only consider a weaker
consistency notion that restricts itself to the bounds of the domains:
\emph{bounds consistency}.

\begin{definition}[Bounds consistency]
  Given $C$ a constraint, $D$ a Cartesian product of domains, and $x$
  a variable in \CtrScope{C}, $x$ is said \emph{bounds
    consistent} w.r.t.\ $C$ and \VarDomain{x} if and only if the
  following property holds:
\begin{equation*}
  \min(\VarDomain{x})\in\CtrProjection{C}{x}{D}\ \wedge\ 
  \max(\VarDomain{x})\in\CtrProjection{C}{x}{D}
\end{equation*}
A constraint $C$ is said bounds consistent w.r.t.\ a Cartesian product
of domains $D$ if and only if every variable $x$ in its scope is
bounds consistent relative to $C$ and \VarDomain{x}. A constraint
system is bounds consistent w.r.t.\ $D$ if and only if each of its
constraints is bounds consistent w.r.t.\ $D$.
\end{definition}

Solving a constraint problem $\Network$ means computing all tuples of
the product of domains that satisfy \emph{all} the constraints. The
general algorithm is a search strategy called \emph{backtracking}. The
computation state is a search tree whose nodes are labelled by a set
of domains. The backtracking algorithm requires a time exponential in
the number of variables in the worst case. Its performances can be
improved with local consistency enforcement to reduce the domains
prior to performing the search. For instance, the domain of a variable
$x$ that is not bounds consistent relative to a constraint $C$ can be
reduced by the following operation:
\vspace{-2mm}
\begin{equation}\label{eq:revisebounds}
\ReviseBounds{C}{D,x} = \Interval{\min(\VarDomain{x}\cap
  \CtrProjection{C}{x}{D})}{\max(\VarDomain{x}\cap\CtrProjection{C}{x}{D})}
\end{equation}
The consistency of a constraint network is obtained by a constraint
propagation algorithm. An AC3-like algorithm~\cite{Mackworth:AI77} is
given in Table~\ref{algo:bounds}, where the
``$\mathrm{\mathop{ReviseBounds}}(C,D)$'' operation applies
$\mathrm{\mathop{ReviseBounds}}$ as given by
Eq.~\eqref{eq:revisebounds} on each variable in \CtrScope{C} in an
arbitrary order. 

\begin{proposition}[Worst-case for BOUNDS-CONSISTENCY~\cite{Lhomme:IJCAI93}]
\label{prop:BC}
Given $m$ the number of constraints to consider, $d$ the size of the
largest domain, $k$ the maximum
number of variables in any constraint, and $\gamma$ the cost of
revising a constraint (that is, the number of projections to apply),
BOUNDS-CONSISTENCY incurs a cost of the order $O(mkd\gamma)$ in the worst case 
to achieve bounds consistency.
\end{proposition}

\begin{table}[htbp]
\vspace{-5mm}
\caption{Computation of a bounds consistent constraint problem}
\label{algo:bounds}
\vspace{-5mm}
\AlgoNumSet{}
\begin{tabbing}
\hspace*{5mm}\=\hspace*{5mm}\=\hspace*{5mm}\=\kill
Algorithm: \AlgoName{BOUNDS-CONSISTENCY}\\
\AlgoKey{Input}: a constraint problem
    $\Network=(\NetSetVar,\NetSetDom,\NetSetCtr)$\\
\AlgoKey{Output}: a bounds consistent equivalent constraint problem\\
\AlgoNum{}\> $\GSet{S} \gets \NetSetCtr$\\
\AlgoNum{}\> \AlgoKey{while} $\GSet{S}\neq\emptyset$ \AlgoKey{do}\\
\AlgoNum{}\> \> $C\gets\text{choose an element of }\GSet{S}$\\
\AlgoNum{}\> \> $D'\gets \ReviseBounds{C}{D}$\\
\AlgoNum{}\> \> \AlgoKey{foreach} $x_i$ s.t.\ $D'\lbrack x_i\rbrack\subsetneq\VarDomain{x_i}$ \AlgoKey{do}\\
\AlgoNum{}\> \> \> $\GSet{S}\gets\GSet{S}\cup\{C'\mid
          C'\in\NetSetCtr\wedge x_i\in\CtrScope{C'}\}$\\
\AlgoNum{}\> \> \> $\VarDomain{x_i}\leftarrow D'\lbrack x_i\rbrack$\\
\AlgoNum{}\> \> \AlgoKey{endfor}\\
\AlgoNum{}\> \> $\GSet{S}\gets\GSet{S}\setminus \{C\}$ \hspace*{1cm}\% $\mathrm{\mathop{ReviseBounds}}$ is idempotent\\
\AlgoNum{}\> \AlgoKey{endwhile}
\end{tabbing}
\vspace{-10mm}
\end{table}


As noted by Dechter~\cite{Dechter:book2003}, it may not be wise to
spend too much time in trying to remove as many inconsistent values as
possible by enforcing a ``perfect arc consistency'' on each constraint
with Alg.~\texttt{BOUNDS-CONSISTENCY}. It may be indeed more efficient
to defer part of the work to the search process.

The amount of work performed can be reduced by adopting the notion of
\emph{directional consistency}~\cite{Dechter-Pearl:AI87},
where inferences are restricted according to a particular variable ordering.

\begin{definition}[Directional bounds consistency]
  Given a Cartesian product of domains $D$, a constraint system $R$ is
  \emph{directional bounds consistent} relative to $D$ and a strict
  partial ordering on variables if and only if for every variable $x$
  and for every constraint $C\in R$ on $x$ such that no variable $y$
  of its scope is smaller than $x$, $x$ is bounds consistent relative
  to $C$ and \VarDomain{x}.
\end{definition}

A propagation algorithm for directional bounds consistency, called
\DBC{}, is presented in Table~\ref{algo:directional}.  It is adapted
from Dechter's directional consistency algorithms.  We introduce a
partition $\Gamma_1,\dots,\Gamma_q$ of the set of variables $V$ that
is compatible with the given partial ordering ``$\prec$'': two
different variables $x$ and $y$, such that $x$ precedes $y$ ($x\prec
y$), must belong to two different sets $\Gamma_i$ and $\Gamma_j$ with
$i<j$. The worst case complexity for Alg.~DBC is $O(km)$ revisions,
with the same notations as in Prop.~\ref{prop:BC}.

\vspace{-4mm}
\section{Directional Bounds Consistency and HC4}
\label{sec:directional-bounds-consistency}

In this section, we first adapt the framework presented in the
previous section to the case of continuous CSPs, and we then show that
the revising procedure for the HC4 algorithm enforces a directional
bounds consistency.

\vspace{-3mm}
\subsection{Bounds Consistency on Continuous Domains}

The definition for bounds consistency has to be slightly adjusted when
one considers continuous domains since we cannot handle real numbers
with perfect exactness on computers. We then restrict ourselves to
real domains from a set \ISet\ represented by intervals whose bounds
are representable numbers (\emph{floating-point numbers}). The size of
a domain is then equal to the number of floating-point numbers it
contains. We also introduce an \emph{approximation function} hull to
manipulate real relations, defined by: $\Hull{\rho} =
\bigcap\{B\in\ISet^n\mid \rho\subseteq B\}$, for all $\rho\subseteq\RSet^n$.

\begin{table}[htbp]
\vspace{-5mm}
\caption{Computation of a directional bounds consistent constraint problem}
\label{algo:directional}
\vspace{-5mm}
\AlgoNumSet{}
\begin{tabbing}
\hspace*{5mm}\=\hspace*{5mm}\=\hspace*{5mm}\=\hspace*{5mm}\=\kill
Algorithm: \AlgoName{DIRECTIONAL-BOUNDS-CONSISTENCY} (\DBC{})\\
\AlgoKey{Input}: \> \>~ --\ a constraint problem
    $\Network=(\NetSetVar,\NetSetDom,\NetSetCtr)$\\
\> \>~    --\ a strict partial ordering $\prec$ over $\NetSetVar$\\
\> \>~    --\ an ordered partition $\Gamma_1,\dots,\Gamma_q$ of $\NetSetVar$ compatible with $\prec$\\
\AlgoKey{Output}: a directional bounds consistent equivalent constraint problem\\
\AlgoNum{}\> \AlgoKey{for} $i=q$ \AlgoKey{downto} $1$ \AlgoKey{do}\\
\AlgoNum{}\> \> \AlgoKey{foreach} $C\in\NetSetCtr$ such that
    $\Gamma_i\subseteq\CtrScope{C}$ \AlgoKey{and}
    $\CtrScope{C}\subseteq\Gamma_1\cup\cdots\cup\Gamma_i$ \AlgoKey{do}\\

\AlgoNum{}\> \> \> \AlgoKey{foreach} $x\in\CtrScope{C}-\Gamma_i$ \AlgoKey{do}\\

\AlgoNum{}\> \> \> \> $\VarDomain{x}\leftarrow\ReviseBounds{C}{D,x}$\\

\AlgoNum{}\> \> \> \AlgoKey{endfor}\\
\AlgoNum{}\> \> \AlgoKey{endfor}\\
\AlgoNum{}\> \AlgoKey{endfor}
\end{tabbing}
\vspace{-8mm}
\end{table}

\noindent The definition for bounds consistency then becomes:

\begin{definition}[Continuous bounds consistency]
\label{def:continuous-bounds-consistency}
Given $C$ a constraint, $D$ a Cartesian product of domains, and $x$ a
variable in \CtrScope{C}, $x$ is said \emph{bounds consistent}
w.r.t.\ $C$ and \VarDomain{x} if and only if the following property holds:
\begin{equation*}
  \min(\VarDomain{x})\in\Hull{\CtrProjection{C}{x}{D}}\ \wedge\ 
  \max(\VarDomain{x})\in\Hull{\CtrProjection{C}{x}{D}}
\end{equation*}
A constraint $C$ is said bounds consistent w.r.t.\ a Cartesian product
of domains $D$ if and only if every variable $x$ in its scope is
bounds consistent relative to $C$ and \VarDomain{x}. A constraint
system is bounds consistent w.r.t.\ $D$ if and only if each of its
constraints is bounds consistent w.r.t.\ $D$.
\end{definition}
As a tribute to legibility, \emph{bounds consistency} will be used
from now on as a shorthand for \emph{continuous bounds consistency}.

\vspace{-3mm}
\subsection{HC3revise: a Revising Procedure for Bounds Consistency}

According to Def.~\ref{def:continuous-bounds-consistency}, the
enforcement of bounds consistency on a real constraint requires the
ability to project it on each of its variables and to intersect the
projections with their domains.

In the general case, the accumulation of rounding errors and the
difficulty to express one variable in terms of the others will
preclude us from computing a precise projection of a constraint.  
However, such a computation may be performed for
constraints involving no more than one operation ($+$, $\times$,
$\cos$, \dots), which corresponds to binary and ternary constraints
such as $x\times y = z$, $\cos(x)=y$, \dots


As a consequence, complex constraints have to be decomposed into
conjunctions of binary and ternary constraints (the \emph{primitives})
prior to the solving process.

Enforcing bounds consistency on a primitive is obtained through the
use of interval arithmetic~\cite{Moore:66}. To be more specific, the
revising procedure for a constraint like $C\colon x+y=z$ is
implemented as follows:
\begin{equation*}
  \NewMathOp{HC3revise}{C,D}=\left\{\begin{array}{ll}
      \ReviseBounds{C}{D,x}\colon 
      &\VarDomain{x} \gets \VarDomain{x}\cap (\VarDomain{z}\ominus\VarDomain{y})\\
      \ReviseBounds{C}{D,y}\colon
      &\VarDomain{y} \gets \VarDomain{y}\cap (\VarDomain{z}\ominus\VarDomain{x})\\
      \ReviseBounds{C}{D,z}\colon
      &\VarDomain{z} \gets \VarDomain{z}\cap (\VarDomain{x}\oplus\VarDomain{y})
    \end{array}\right.
\end{equation*}
\vspace{-5mm}

\noindent where $\ominus$ and $\oplus$ are interval extensions of the corresponding real arithmetic
operators.

Enforcing bounds consistency on a constraint system is performed in
two steps: the original system is first decomposed into a conjunction
of primitives, adding fresh variables in the process; the new system
of primitives is then handled with the BOUNDS-CONSISTENCY algorithm
described in Table~\ref{algo:bounds}, where the ReviseBounds procedure
is performed by an HC3revise algorithm for each primitive.

\vspace{-4mm}
\subsection{HC4revise: a Revising Procedure for Directional Bounds Consistency}

\vspace{-2mm}
The HC4 algorithm was originally presented by its
authors~\cite{Benhamou:ICLP1999} as an efficient means to compute
bounds consistency on complex constraints with no variable occurring
more than once (called \emph{admissible constraints} in the rest of
the paper). It was demonstrated to be still more efficient than HC3 to
solve constraints with variables occurring several times, though it
was not clear at the time what consistency property is enforced on any
single constraint in that case.

To answer that question, we first describe briefly below the revising
procedure HC4revise of HC4 for one constraint $C$ as it was originally
presented, that is in terms of a two sweeps procedure over the
expression tree of $C$. We will then relate this algorithm to the ones presented
in Section~\ref{sec:discrete-csp}.

To keep the presentation short, the HC4revise algorithm will be
described by means of a simple example. The reader is referred to the
original paper~\cite{Benhamou:ICLP1999} for an extended description.

\begin{figure}[htbp]
\vspace{-8mm}
  \begin{center}
    \subfigure[Forward sweep]{\label{fig:hc4-forward-sweep}%
      \includegraphics[bb=31 327 567 537,width=.48\textwidth]{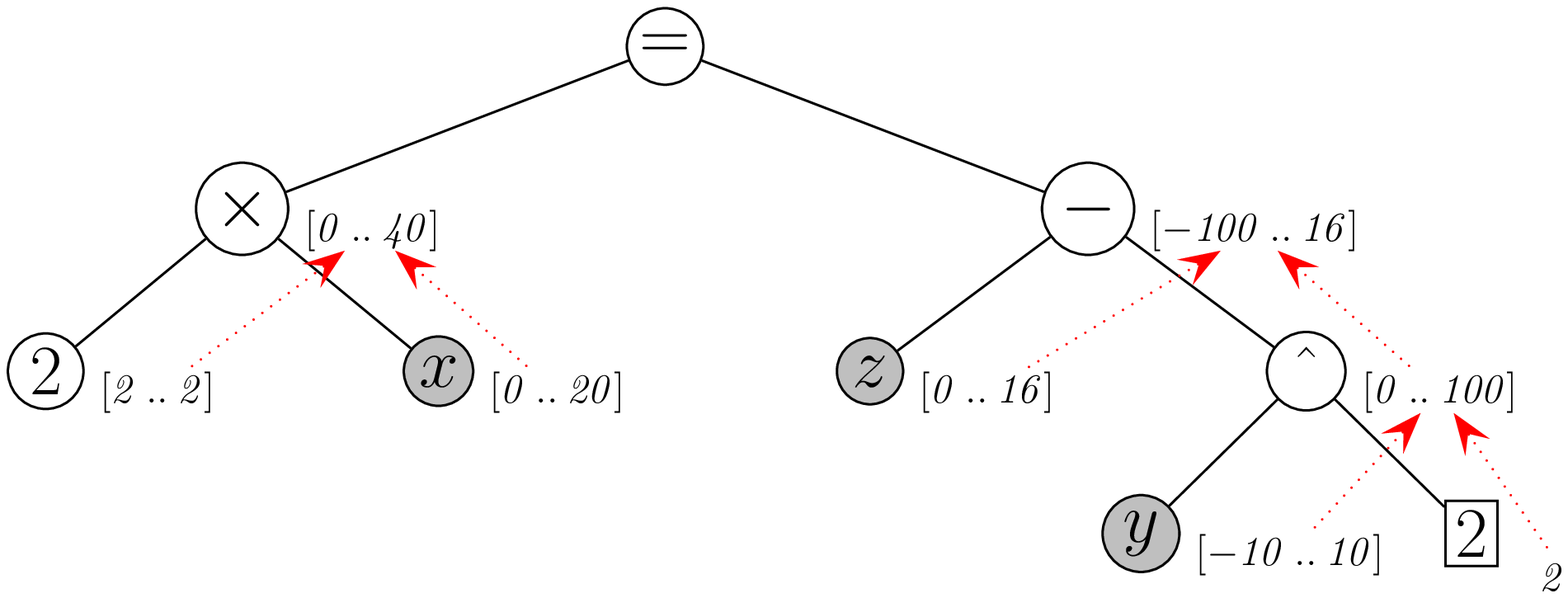}}
    \subfigure[Backward sweep]{\label{fig:hc4-backward-sweep}%
      \includegraphics[bb=31 329 565 548,width=.48\textwidth]{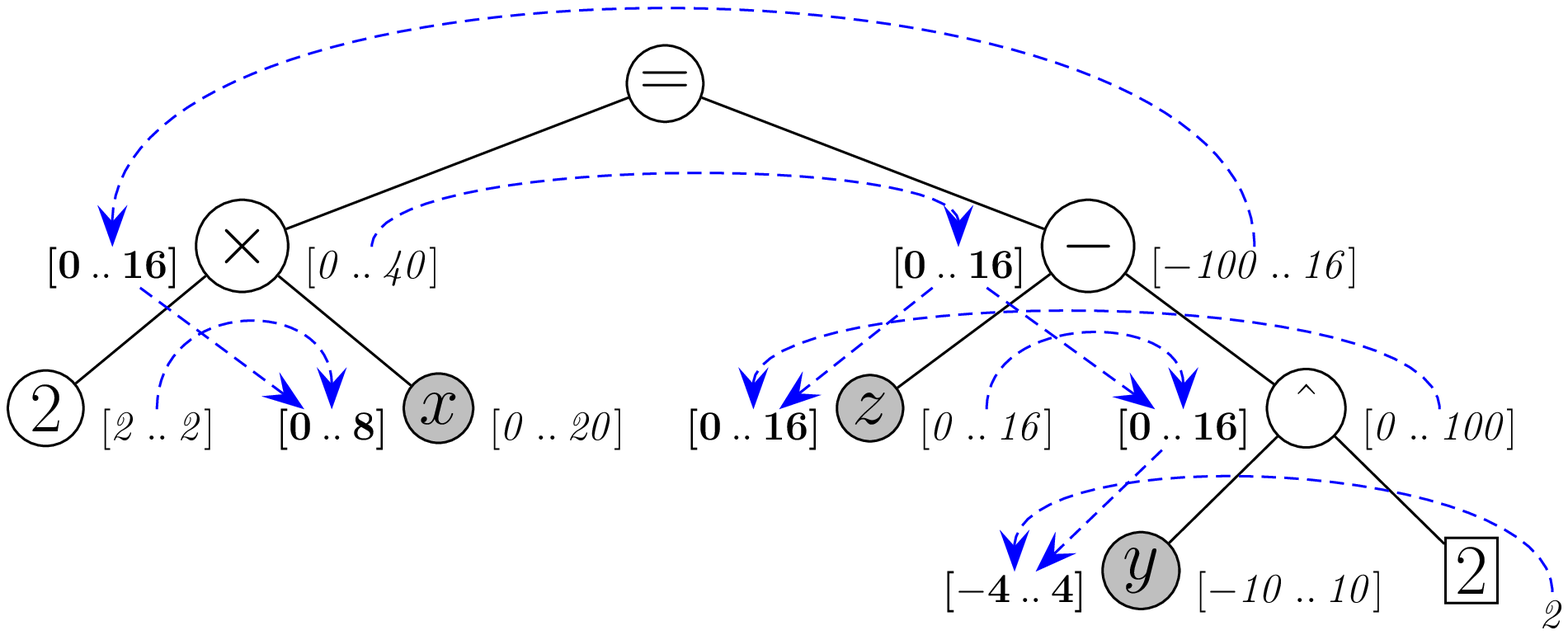}}
    \vspace{-4mm}
    \caption{HC4revise on the constraint $2x=z-y^2$}
    \label{fig:HC4revise}
  \end{center}
\vspace{-8mm}
\end{figure}

Given the constraint $C\colon 2x=z-y^2$, HC4revise first evaluates the
left-hand and right-hand parts of the equation using interval
arithmetic, saving at each node the result of the local evaluation
(see Fig.~\ref{fig:hc4-forward-sweep}). In a second sweep from top to
bottom on the expression tree (see Fig.~\ref{fig:hc4-backward-sweep}),
the domains computed during the first bottom-up sweep are used to
project the relation at each node on the remaining variables.

Given the constraint set $\Delta_C= \{2x=\alpha_1, y^2=\alpha_2,
z-\alpha_2=\alpha_3, \alpha_1=\alpha_3\}$ obtained by decomposing $C$
into primitives, it is straightforward to show that HC4revise simply
applies all the ReviseBounds procedures in $\Delta_C$ in a specific 
order---induced by the expression tree of $C$---noted $\omega_C$ in the sequel.

To be more specific, HC4revise first applies the ReviseBounds
procedure for the right-hand variable of all the primitives up to the
root, and then the ReviseBounds procedures for the left-hand
variables:
\vspace{-3mm}
\begin{multline*}
  \NewMathOp{HC4revise}{C,D}=\\
  \hspace*{4mm}\begin{array}{lllll}
  1.&\text{ReviseBounds}(2x=\alpha_1,D,\alpha_1) & \text{\hbox to 1mm{}}& 
  6.&\text{ReviseBounds}(2x=\alpha_1,D,x)\\
  2.&\text{ReviseBounds}(y^2=\alpha_2,D,\alpha_2)&& 
  7.&\text{ReviseBounds}(z-\alpha_2=\alpha_3,D,z)\\
  3.&\text{ReviseBounds}(z-\alpha_2=\alpha_3,D,\alpha_3) && 
  8.&\text{ReviseBounds}(z-\alpha_2=\alpha_3,D,\alpha_2)\\
  4.&\text{ReviseBounds}(\alpha_1=\alpha_3,D,\alpha_3) && 
  9.&\text{ReviseBounds}(y^2=\alpha_2,D,y)\\
  5.&\text{ReviseBounds}(\alpha_1=\alpha_3,D,\alpha_1)\\
  \end{array}
\end{multline*}
Note that, so doing, the domain of each fresh
variable introduced by the decomposition process is set to a useful
value before being used in the computation of the domain of any other
variable.

For admissible constraints, the HC4revise algorithm can be implemented
using the DBC algorithm by considering two well-chosen partitions of
the set of variables of the decomposed problem. Non-admissible
constraints need being made admissible by adding new variables to
replace multiple occurrences. The partitioning scheme is given by tree
traversals as follows: The first partition $\Gamma$ is obtained by a
right-to-left preorder traversal of the tree where visiting a node has
the side effect of computing the set of variables associated with its
children. The underlying strict partial ordering of the variables is
such that a child is greater than its parent. The second partition
$\Gamma'$ is obtained by inverting the partition computed by a
left-to-right preorder traversal of the tree where the visit of a root
node associated with a variable $x$ just computes the set $\{x\}$.
The underlying strict partial ordering is such that a child is smaller
than its parent.  HC4revise is equivalent to applying DBC on $\Gamma$
and then on $\Gamma'$.

Going back to our example, let us consider the CSP $P=(V,D,\Delta_C)$, with
$V=\{x,y,z,\alpha_1,\alpha_2,\alpha_3\}$,
$D=\VarDomain{x}\times\VarDomain{y}\times\VarDomain{z}\times\VarDomain{\alpha_1}
\times\VarDomain{\alpha_2}\times\VarDomain{\alpha_3}$, and $\Delta_C$ defined
as above. Let us also consider a dummy fresh variable $\alpha_0$
supposed to be in the scope of the constraint represented by the root
node and its children ($\alpha_1=\alpha_3$), which is only introduced
to initialize the computation of projections\footnote{Alternatively,
  the constraint $\alpha_1=\alpha_3$ could be replaced by the
  equivalent one $\alpha_1-\alpha_3=\alpha_0$, with $\alpha_0$
  constrained to be equal to $0$.}. The partitions used to apply
HC4revise on $P$ by using Alg.~DBC are then as follows:
\begin{equation*}
\left\lbrace
\begin{array}{rcl}
  \Gamma  & = & \{\alpha_0\}, \{\alpha_1,\alpha_3\},\{z,\alpha_2\},\{y\},\{x\}\\
  \Gamma' & = & \{y\},\{\alpha_2\},\{z\},\{\alpha_3\},\{x\},\{\alpha_1\},\{\alpha_0\}\\
\end{array}
\right.
\end{equation*}

Let $\gamma_C$ (resp.\ $\gamma'_C$) be the partial ordering induced by
$\Gamma$ (resp.\ $\Gamma'$) on the variables.  With its two sweeps on
a tree-shaped constraint network, HC4revise appears very similar to
Dechter's \emph{ADAPTIVE-TREE-CONSISTENCY} algorithm~\cite[p.\ 
265]{Dechter:book2003}.  More importantly, the constraint network
processed by Alg. DBC being a tree, we can state a result
analogous to the one stated by Freuder~\cite{Freuder:JACM82} for arc
consistency, which says that, on tree-shaped constraint networks, a
bottom-up sweep followed by a top-down sweep are all it takes to
enforce bounds consistency:

\begin{proposition}[Consistency enforced by HC4revise]
\label{prop:hc4revise}
Given $C$ a constraint and $D$ a Cartesian product of domains for the
variables in \CtrScope{C}, let $\Delta_C$ be the set of primitive
constraints obtained by decomposing $C$. We have:
  \begin{enumerate}
  \item $\Delta_C$ is directional bounds consistent w.r.t.\ $\gamma'_C$ and
    $D''=\NewMathOp{HC4revise}{C,D}$;
  \item if $C$ is an admissible constraint, the constraint system represented by 
    $\Delta_C$ is bounds consistent w.r.t.\ $D''=\NewMathOp{HC4revise}{C,D}$.
  \end{enumerate}
  \vspace{-4mm}
  \begin{proof}
    Let $\Gamma$ and $\Gamma'$ be two partitions for the variables in 
    $V=\bigcup_{C'\in \Delta_C} \CtrScope{C'}$ defined as described above. 
    As stated previously, we
    have $\NewMathOp{HC4revise}{C,D}=\NewMathOp{DBC}{\langle V,D',\Delta_C\rangle,
      \gamma'_C,\Gamma'}$, 
    where $D'=\NewMathOp{DBC}{\langle V,D,\Delta_C\rangle,\gamma_C,\Gamma}$.
    
    The first point follows directly from this identity. To prove the
    second point, let us consider the set $\Pi_C$ of projection
    operators implementing the ReviseBounds procedures for the
    primitives in $\Delta_C$. The HC3 algorithm applied on $\Delta_C$ and
    $D$ would compute the greatest common fixed-point $\top$ included in
    $D$ of these operators, which is unique since they all are
    monotonous~\cite{Benhamou-Older:JLP97}. By design of HC3, $\Delta_C$ 
    is bounds consistent w.r.t.\ $\top$.
    
    Consider now HC4revise called on $C$ and $D$, which applies each
    of the operators in $\Pi_C$ once in the order $\omega_C$: 
    \begin{itemize}
      \item either it computes a fixed-point of $\Pi_C$, which must be the
        greatest fixed-point $\top$, by unicity of the gfp and by contractance
        of the operators in $\Pi_C$,
      \item or, it is possible to narrow further the domains of the
        variables by applying one of the operators in $\Pi_C$. Let
        $\pi_1\colon \beta_1\gets f_1(\beta_2,\dots,\beta_k)$ be this
        operator.  Consider the case where $\pi_1$ is an operator
        applied during the bottom-up sweep. According to the order
        $\omega_C$, $\pi_2\colon \beta_2\gets
        f_2(\beta_1,\beta_3,\dots,\beta_k), \dots,\allowbreak 
        \pi_k\colon\beta_k\gets f_k(\beta_1,\dots,\beta_{k-1})$ have then been applied
        during the top-down sweep, that is, after having applied
        $\pi_1$. The constraint $C$ being admissible by hypothesis,
        each variable occurs in only one node in the tree, and then,
        $\beta_2,\dots,\beta_k$ cannot have been modified after
        having applied $\pi_2, \dots, \pi_k$. Consequently, reapplying
        $\pi_1$ after the two sweeps cannot narrow down $\beta_1$
        further, since its most up-to-date value has already been used
        to compute the current domains for $\beta_2,\dots,\beta_k$ and $\pi_1$ is 
        idempotent~\cite{Benhamou-Older:JLP97}. We
        may then conclude that no operator applied during the
        bottom-up sweep needs to be reapplied. We can use the same
        arguments for an operator $\pi_1$ that was first applied
        during the top-down sweep.

        As a consequence, no operator in $\Pi_C$ needs being reapplied,
        which contradict our hypothesis that we had not reached a
        fixed-point. As said above, if HC4revise computes a fixed-point,
        it is necessarily the greatest common fixed-point $\top$
        included in $D$ of the operators in $\Pi_C$, and then bounds
        consistency has been enforced on $\Delta_C$.\qed
      \end{itemize}
  \end{proof}
\end{proposition}

\section{Theoretical Analysis of HC4 vs.\ HC3}
\label{sec:theoretical-analysis}

The experimental results given in Benhamou et al.'s
paper~\cite{Benhamou:ICLP1999} as well as in
Section~\ref{sec:experimental-results} below clearly exhibit
the superiority of HC4 versus HC3 to solve large constraint problems.
We present in this section the theoretical analysis of these two
algorithms.

\subsection{Applying HC3 and HC4 to One Constraint Only}
\label{sec:complexity-one-constraint}

In this section as well as in the next, we will consider the
projection of a primitive constraint onto a variable as the atomic
instruction whose count will serve to characterize the efficiency of
the algorithms analyzed.

Let us determine the number of projections to apply in the worst case
to enforce bounds consistency on a single admissible constraint (not
necessarily primitive). Given a constraint $C$ and its decomposition
$\Delta_C$ into $p$ primitives, let $k$ be the maximum arity of $\Delta_C$ defined by
$k=\max_{C'\in \Delta_C}\text{arity}(C')$. 

Let $o$ be the number of nodes in the expression tree of $C$. It is easy to observe that
$o$ and $p$ are of the same order (more precisely, we have $o/p \sim k$).

\begin{proposition}[Worst-case for HC4 on one constraint]
\label{prop:hc4revise-one}
HC4 enforces bounds consistency on the system $\Delta_C$ of
constraints originating from an admissible constraint $C$ that is
decomposable into $p$ primitives in $O(p)$
projections.
\begin{proof}
  As stressed in the previous section, the tree-shaped constraint
  network composed naturally by the constraints in $\Delta_C$ implies that
  HC4 will enforce bounds consistency on $\Delta_C$ once its two
  sweeps complete. The number of projections applied is then equal to
  the $p$ evaluations during the forward sweep plus the projections on
  the remaining variables for each primitive constraint during the
  backward sweep, that is at most $(k-1)p$. Overall, the number of projections
  for HC4revise to enforce bounds consistency is then at most $p+(k-1)p=kp$
  projections. \qed
\end{proof}
\end{proposition}

\begin{proposition}[Worst-case for HC3 on one constraint]
\label{prop:hc3-one}
  HC3 enforces bounds consistency on the system $\Delta_C$ of
constraints originating from an admissible constraint $C$ that
  is decomposable into $p$ primitives in $O(p^2)$ projections.
   \begin{proof}
     From Prop.~\ref{prop:hc4revise}, we know that bounds consistency
     is obtained when the information represented by the domain of
     each variable (both the user's ones as well as the fresh ones
     introduced by the decomposition process) is passed to all the
     other variables in the tree of $C$.  An efficient way to do that
     indeed corresponds to Alg.\ HC4revise.  Since the tree contains at most
     $o$ variables, there are at most $o(o-1)$ informations to
     exchange.  Considering an algorithm like BOUNDS-CONSISTENCY, each
     time a primitive is considered, at least one information is
     transfered from one variable to the others in its scope (which
     does not imply necessarily any modification in the associated
     domains). We then obtain at most $o(o-1)$ calls to primitives.
     Using the fact that $o$ and $p$ are related by $k$, and that each
     primitive requires applying at most $k$ projections, with $k$ a
     constant, the result follows.\qed
  \end{proof}
\end{proposition}

Relating the worst-cases for HC3 and HC4, we obtain that the ratio
$\text{HC3}/\text{HC4}$ is of order $p$ for one constraint in the worst case.

\subsection{Applying HC3 and HC4 on a Constraint System}
\label{sec:complexity-constraint-system}

Given a system $R$ of $m$ admissible constraints on $n$ variables
$x_1,\dots,x_n$, let $d$ be the size of the largest initial domain.
Given $\Delta_R$ the constraint system
obtained from decomposing the constraints in $R$ into primitives, let $k$
be the maximum arity of $\Delta_R$.

As said before, Alg.\ HC4 enforces bounds consistency on $\Delta_R$ by
applying Alg.\ DIRECTIONAL-BOUNDS-CONSISTENCY on each constraint in
$R$ twice every time.  We stress again that bounds consistency is
eventually computed only because we consider \emph{admissible}
constraints in $R$, that is, constraints containing no variable
occurring more than once.

\begin{proposition}[Worst-case for HC4]
\label{prop:HC4}
The number of projections to apply to achieve bounds consistency with
HC4 on a constraint system $\Delta_R$ obtained from a set $R$ of $m$
admissible constraints on $n$ variables is of the order $O(mndp)$ in
the worst case.
\begin{proof}
  The number of constraints to consider is $m$; as shown in
  Prop.~\ref{prop:hc4revise-one}, the cost to apply
  HC4revise is of the order $O(p)$, with $p$ the maximum number of
  primitives necessary to decompose each constraint in $R$.  The
  result then follows directly from Prop.~\ref{prop:BC}.  \qed
\end{proof}
\end{proposition}

\begin{proposition}[Worst-case for HC3]
\label{prop:HC3}
The number of projections to apply to achieve bounds consistency with
HC3 on a constraint system $\Delta_R$ obtained from a set $R$ of $m$
admissible constraints on $n$ variables is of the order $O(mpd)$ in
the worst case.
 \begin{proof}
   First, we must note that the number of constraint to consider is no
   longer $m$ but at most $mp$, since each constraint has to be
   decomposed beforehand. The maximum cost of applying HC3revise on a
   constraint is $k$ (we apply each projection once).  Each constraint
   can only be reinvoked at most $kd$ times.  Using
   Prop.~\ref{prop:BC}, we obtain a worst-case estimate of $mpkdk$,
   with $k$ a constant.  \qed
 \end{proof}
\end{proposition}

Relating the costs of computing bounds consistency with either HC4 or HC3, we
now obtain $HC3/HC4 \sim mpd/(mndp)$, that is, $HC3/HC4 \sim 1/n$, 
which means that the relation is now inverted compared with the case of one 
admissible constraint only. We then have that the ratio $HC4/HC3$ is
of the order of the number of variables in the problem in the
worst-case.  As we will see in the next section, this pessimistic
result is contradicted by all experimental results. It is however easy
to get an intuitive understanding of it if one considers that, in the
worst case, HC4revise may be called for a constraint each time only
one value is removed from the domains of the variables in its scope.
Using HC4revise leads to considering $p$ times less constraints than
with HC3revise, since the original constraints do not have to be
decomposed.  However, HC4revise is $p$ times more costly to apply than
HC3revise. Overall, HC4 is then penalized by the number of
opportunities to reinvoke HC4revise, of the order $nd$ (vs.\ $kd$ for
HC3revise in HC3). Note also that, considering Prop.~\ref{prop:HC3}
for a system of only one constraint, we obtain that the number of
projections to apply is of the order $O(pd)$ vs.\ $O(p^2)$ if we
consider Prop.~\ref{prop:hc3-one}. This contradiction is only apparent
since, if $d<p$, the $O(p^2)$ result is clearly pessimistic since it
is not possible to apply $p^2$ projections (there are not enough
values to discard overall), and if $d>p$, $O(pd)$ is pessimistic since
at most $p^2$ calls suffice to broadcast the information contained by each
node in the tree-shaped network of the constraint.

\vspace{-4mm}
\section{Experimental Results}\label{sec:experimental-results}

We present the results of both HC4 and HC3 on four standard benchmarks
from the interval constraint community.  They were chosen so as to be
scalable at will and to exhibit various behaviours of the algorithms.
As a side note, it is important to remember that these algorithms are 
often outperformed by other algorithms. Their study is still
pertinent, however, since they serve as basic procedures in these
more efficient algorithms.

It is important to note also that, originally, none of these problems is
admissible. In order to show the impact of admissibility, we have
factorized the constraints of one of them.

All the problems have been solved on an AMD Athlon 900~MHz under Linux,
using a C++ interval constraint library written for our tests based on
the gaol\footnote{Interval C++ library available at
  \url{http://sf.net/projects/gaol/}} interval arithmetic library.  In
order to avoid any interference, no optimization (e.g.,
\emph{improvement factor}) was used.

For each benchmark, four different methods have been used:
\begin{itemize}
\item \texttt{HC3}, which enforces bounds consistency on the decomposed system;
\item \texttt{HC3sb}, which uses \emph{S-boxes}~\cite{Goualard:ASE99}:
  each user constraint $C$ is decomposed into a separate set of
  primitives and gives rise to a \ReviseBoundsZA\ procedure $RB_C$
  that enforces bounds consistency on this set by using HC3revise
  procedures for each primitive, and propagating modifications with
  Alg.~\ref{algo:bounds}. All the $RB_C$ methods for the constraints in
  the user system are then handled themselves by
  Alg.~\ref{algo:bounds}. This propagation scheme forces consistency to be
  enforced locally for each user constraint before reconsidering the others;
\item \texttt{HC4}, which enforces a directional bounds consistency
  (and not bounds consistency, since the constraints are not
  admissible) on each constraint using HC4revise, and which uses
  Alg.~\ref{algo:bounds} for the propagation over the constraints in
  the system;
\item \texttt{HC4sb}, which uses one S-box per user constraint. As a consequence, HC4revise
  is called as many times as necessary to reach a fixed-point for any non-admissible constraint.
\end{itemize}

\begin{figure}[htbp]
  \vspace{-8mm}
  \begin{center}
    \includegraphics[width=.95\textwidth]{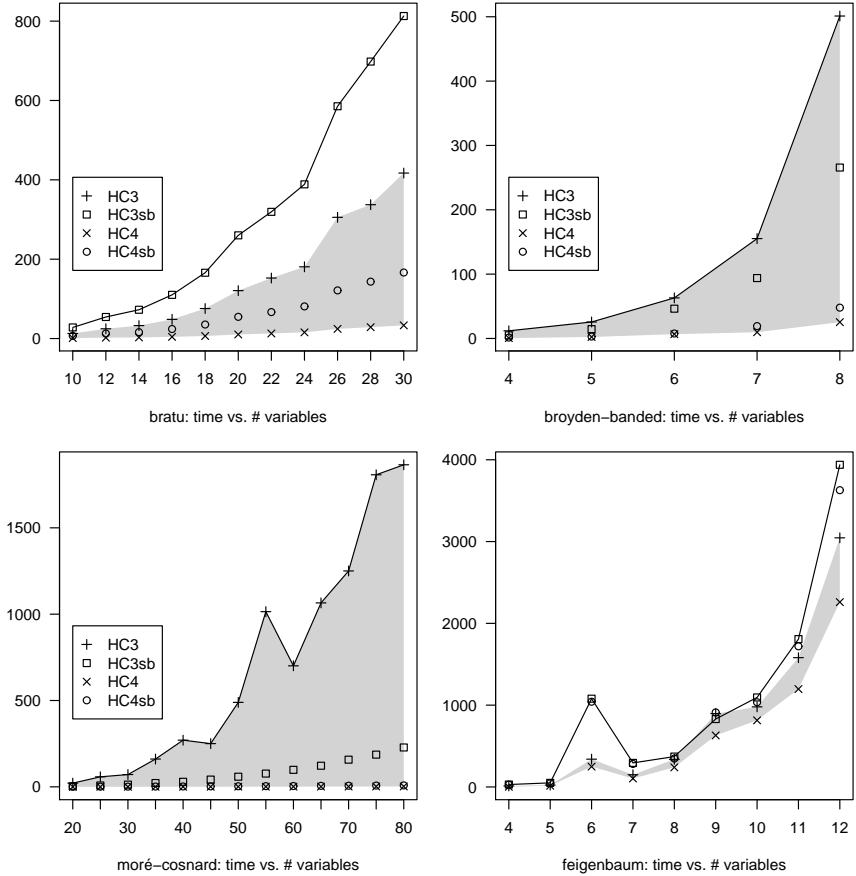}
    \vspace{-3mm}
    \caption{Solving times in seconds for all benchmarks}
    \label{fig:problems}
  \end{center}
\vspace{-8mm}
\end{figure}

Each graphics provided (see Fig.~\ref{fig:problems}) displays the
computation time in seconds required to find all solutions up to an
accuracy of $10^{-8}$ (difference between the lower and upper bounds
of the intervals) for each method.

The \texttt{bratu} constraint system modelizes a problem in combustion
theory. It is a square, sparse and quasi-linear system of $n+2$
equations:

\begin{equation*}
  \left\{\begin{array}{l}
    \forall k\in\{1,\dots,n\}\colon x_{k-1}-2x_k+x_{k+1}+\exp(x_k)/(n+1)^2=0,\\
    x_0=x_{n+1}=0,\\
    \forall i\in\{1,\dots,n\}\colon x_i\in[-10^8,10^8]
  \end{array}\right.
\end{equation*}
The largest number of nodes per constraint is independent of the size
of the problem and is equal to 12 in our implementation.

As already reported by Benhamou et al.~\cite{Benhamou:ICLP1999}, HC4
appears more efficient than HC3 to solve all instances of the problem,
and its advantage grows with their size. Localizing the propagation
does not seem a good strategy here, since HC3sb and HC4sb both perform
poorly in terms of the number of projections computed\footnote{Due to
  lack of space, all the graphics corresponding to the number of
  projections have been omitted. They are available at the url given
  at the end of Section~\ref{sec:discussion}.}. Interestingly enough,
HC4sb is faster than HC3 while it requires more projections. The first
reason for this discrepancy that comes to mind is that the
``anarchic'' propagation in HC3 has a cost much higher in terms of
management of the set of constraints to reinvoke than the controlled
propagation achieved with HC4sb (see below for another analysis).

The \texttt{broyden-banded} problem is very difficult to solve with
HC3, so that we could only consider small instances of it:
\begin{equation*}
\begin{array}{ll}
   \forall k\in\{1,\dots,n\}\colon 
   &x_k(2+5x_k^2) + 1 - \sum_{j\in J_k} x_j(1+x_j) = 0\\
  & \text{with } J_k = \{j\mid j\neq k\ \wedge \max(1,i-5)\leq j\leq \min(n,i+1)\},\\
  & x_k \in [-10^8,+10^8]\\
\end{array}
\end{equation*}

Contrary to \texttt{bratu}, the number of nodes in the constraints is
not independent of the size of the problem. It follows however a
simple pattern and it is bounded from below by $16$ and from above by
$46$. 

As with \texttt{bratu}, the efficiency of HC4 compared to HC3 is
striking, even on the small number of instances considered. Note that,
here, HC3sb is better than HC3. On the other hand, HC4 is still better
than HC4sb.

The \texttt{mor\'e-cosnard} problem is a nonlinear system obtained
from the discretization of a nonlinear integral equation:
\begin{equation*}
  \forall k\in\{1,\dots,n\}\colon
  \left\{\begin{array}{l}
      x_k\in[-10^8,0],\\
      x_k+\frac{1}{2}[(1-t_k)\sum_{j=1}^k t_j(x_j+t_j+1)^3\\
      \text{\hbox to 1cm{}} +t_k\sum_{j=k+1}^n (1-t_j)(x_j+t_j+1)^2] = 0
      \end{array}\right.
\end{equation*}

The largest number of nodes per constraint grows linearly with the number of
variables in the problem.

HC4 allows to solve this problem up to 1000 times faster than HC3 on
the instances we tested. An original aspect of this benchmark is that
localizing the propagation by using S-boxes seems a good strategy:
HC3sb solves all instances almost as fast as HC4 (see the analysis in
the next section). Note that, once again, though the number of
projections required for HC3sb is almost equal to the one for HC4,
there is still a sizable difference in solving time, which again might
be explained by higher propagation costs in HC3sb.

Lastly, the \texttt{Feigenbaum} problem is a quadratic system:
\vspace{-3mm}
\begin{equation*}
  \left\{\begin{array}{l}
       \forall k\in\{1,\dots,n\}\colon x_k\in[0,100],\\
      \forall k\in\{1,\dots,n-1\}\colon -3.84x_k^2+3.84x_k-x_{k+1} = 0,\\
      -3.84x^2_n+3.84x_n - x_1 = 0
    \end{array}\right.
\end{equation*}

The largest number of nodes per constraint is independent of the size of the problem. It is
equal to 10 in our implementation.

The advantage of HC4 over HC3 is not so striking on this problem. HC4sb and HC3sb do not
fare well either, at least if we consider the computation time.

Parenthetically, the equations in the \texttt{feigenbaum} problem can
easily be factorized so that the resulting problem is only composed of
admissible constraints. Due to lack of space, the corresponding
graphics is not presented here; however, we note that the solving time
is reduced by a factor of more than 500 compared to the original
version.

\section{Discussion}\label{sec:discussion}

Benhamou et al.~\cite{Benhamou:ICLP1999} have tested the HC4 algorithm
on many standard benchmarks. They have shown on each of them the
superiority of HC4 over HC3.  From the results presented in the
preceding section, we have to draw the same conclusions, and to reject
entirely the pessimistic view conveyed by our theoretical analysis.

\begin{figure}[htbp]
\vspace{-9mm}
  \begin{center}
    \includegraphics[width=.75\textwidth]{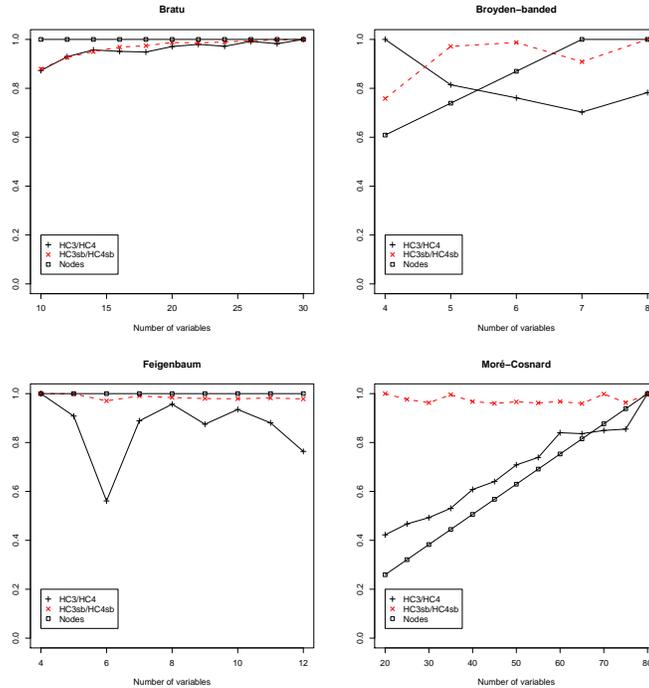}
  \end{center}
  \vspace{-7mm}
  \caption{Impact of the number of nodes per constraint on the number of projections}
  \label{fig:ratio-projections}
  \vspace{-4mm}
\end{figure}

To sum up what has been observed in
Section~\ref{sec:experimental-results}, it appears that it is more
efficient to deal locally with the modification of a domain induced by
some primitive by first reinvoking the other primitives coming from
the decomposition of the same user constraint in problems with large
constraints like \texttt{mor\'e-cosnard}, while the opposite is true
with systems of small constraints such as \texttt{feigenbaum} or
\texttt{bratu}. An intuitive understanding of that may be that the
information just obtained by a reduction is spread and lost in a large
network of primitives while it could be efficiently used locally to
compute projections on a user constraint.
Figure~\ref{fig:ratio-projections} relates the ratio of the number of
projections required by HC3 and HC4, and by HC3sb and HC4sb to the
number of nodes in a constraint. As one may see, the ratio is roughly
constant for HC3 and HC4 when the number of nodes is independent of
the size of the problem, while it increases sharply when the number of
nodes increases with the size of the problem (e.g., 
\texttt{mor\'e-cosnard}). On the other hand, the ratio between HC3sb
and HC4sb stays constant for all problems, a fact particularly
striking with \texttt{mor\'e-cosnard}. It seems a solid evidence that
\emph{localization of the information as obtained from using HC4 (or, to a
lesser extent, HC3sb), is a winning strategy the larger the
constraints in a problem are}.

Note however that HC4 is always more efficient than HC4sb on all the
benchmarks considered. This is consistent with facts long known by
numerical analysts: we show in a paper to come that HC4 may be
considered as a free-steering nonlinear Gauss-Seidel procedure where the inner
iteration is obtained as in the linear case. For this class of
methods, it has been proved experimentally in the past that it is
counterproductive to try to reach some fixed-point in the inner
iteration.

Benhamou et al.\ have shown that one successful strategy to solve
difficult problems is to make HC4revise cooperate with the revise
procedure used to enforce box consistency. A promising direction for
future researches is to investigate other cooperation schemes based on
the analysis of the structure of the constraints (linear, quadratic,
polynomial, \dots) and of the constraint system (full, banded, \dots),
using the cooperation framework presented by Granvilliers and
Monfroy~\cite{GranvilliersMonfroyICLP2003} as a basis.

For the interested reader, all the data used to prepare the figures in
Section~\ref{sec:experimental-results} and many more are available in
tabulated text format at
\url{http://www.sciences.univ-nantes.fr/info/perso/permanents/goualard/dbc-data/}.

\section*{Acknowledgements}

Some results in Section~\ref{sec:complexity-one-constraint}
benefited from discussions with Guillaume Fertin, who pointed out
to us an interesting similarity between the propagation in trees and the
\emph{gossiping problem} in a network.

\vspace{-3mm}

\bibliographystyle{plain}
\bibliography{dbc}

\end{document}